\documentclass[letterpaper, 10 pt, journal, twoside]{IEEEtran}
\ifCLASSINFOpdf
\else
\fi
\usepackage[top=19.1mm, bottom=19.1mm, left=19.1mm, right=19.1mm]{geometry}
\usepackage{lipsum}
\usepackage{colortbl}
\usepackage{url}
\usepackage{graphicx}
\usepackage{amsfonts}
\usepackage{soul} 
\usepackage{color, xcolor} 
\usepackage{amsmath}
\usepackage{amssymb}
\usepackage{multirow}
\usepackage{array}
\usepackage{algorithm}
\usepackage{algpseudocode}
\usepackage{xcolor}

\makeatletter
\algnewcommand{\LineComment}[1]{%
  \Statex \hskip\ALG@thistlm
  \parbox[t]{\dimexpr\linewidth-\ALG@thistlm\relax}{%
    \setlength{\baselineskip}{1.2\baselineskip}%
    \textit{\color{blue!55!black}#1}%
  }%
  \vspace{2pt}%
}
\makeatother

\usepackage{booktabs}
\usepackage[noadjust]{cite}
\usepackage{soul}
\usepackage[colorlinks,allcolors=black]{hyperref}
\usepackage{color}
\ifCLASSOPTIONcompsoc
\usepackage[caption=false, font=normalsize, labelfont=sf, textfont=sf]{subfig}
\else
\usepackage[caption=false, font=footnotesize]{subfig}
\fi

\usepackage{threeparttable}
\usepackage{arydshln}

\begin{document}

%


\title{Do We Really Need Immediate Resets? Rethinking Collision Handling for Efficient Robot Navigation}


\author{Shanze Wang$^{1,2,*}$,
        Xinming Zhang$^{1,4,*}$,
        Siwei Cheng$^{1,4}$,
        Xianghui Wang$^{1,5}$,
        Changwen Chen$^{3}$,
        Hailong Huang$^{2}$,
        and Wei Zhang$^{1,\dagger}$%
\thanks{This work was supported in part by the National Natural Science Foundation of China under Grant 62503251.}%
\thanks{$^{1}$The authors are with the College of Information Science and Technology,
Eastern Institute of Technology, Ningbo, China.
{\tt\footnotesize szwang@eitech.edu.cn, xhwang@eitech.edu.cn,
zhw@eitech.edu.cn, \{xm\_zhang, chengsiwei\}@mail.ustc.edu.cn}}%
\thanks{$^{2}$The authors are with the Department of Aeronautical and Aviation Engineering,
The Hong Kong Polytechnic University, Hung Hom, Kowloon, Hong Kong.
{\tt\footnotesize shanze.wang@connect.polyu.hk, hailong.huang@polyu.edu.hk}}%
\thanks{$^{3}$The authors are with the Department of Computing,
The Hong Kong Polytechnic University, Hung Hom, Kowloon, Hong Kong.
{\tt\footnotesize changwen.chen@polyu.edu.hk}}%
\thanks{$^{4}$The authors are with the School of Computer Science and Technology,
University of Science and Technology of China, Hefei, China.
{\tt\footnotesize \{xm\_zhang, chengsiwei\}@mail.ustc.edu.cn}}%
\thanks{$^{5}$The author is with the Department of Mechanical Engineering,
The Hong Kong Polytechnic University, Hung Hom, Kowloon, Hong Kong.
{\tt\footnotesize carlos.wang@connect.polyu.hk}}%
\thanks{$^{*}$Shanze Wang and Xinming Zhang contributed equally to this work.}%
\thanks{$^{\dagger}$Corresponding author: Wei Zhang.}%
}


%
%

\maketitle
\pagestyle{empty}  
\thispagestyle{empty} 
\begin{abstract} 
Should a single collision necessarily terminate an entire navigation episode? In most deep reinforcement learning (DRL) frameworks for robot navigation, this remains the standard practice: every collision immediately triggers a global environment reset and is penalized as a complete task failure. While a collision during deployment naturally indicates task failure, applying the same treatment during training prevents the agent from exploring challenging obstacle configurations, which slows learning progress in the early training phase. In this work, we challenge this convention and propose a Multi-Collision reset Budget (MCB) framework that decouples local collision termination from global environment resets, allowing the agent to retry difficult configurations within the same episode. 
Simulation experiments show that MCB improves early-stage learning efficiency by reaching target success-rate levels with fewer interactions, with small collision budgets producing the most consistent gains. Real-world experiments on heterogeneous robot platforms further validate the deployability of the learned policies in cluttered environments.
\end{abstract}

%
\IEEEpeerreviewmaketitle

\section{Introduction}
Deep reinforcement learning (DRL) has emerged as a promising paradigm for robot navigation, enabling agents to learn end-to-end policies that directly map raw sensory observations to velocity commands without requiring explicit path planning \cite{Tai2017Virtual}. Through trial-and-error interaction with the environment, DRL agents can acquire reactive collision-avoidance behaviors that generalize across diverse obstacle configurations, and can adapt to changing surroundings based on current observations rather than relying on a pre-built map \cite{Xiao2022Survey}. DRL-based navigation has been progressively extended from structured indoor environments \cite{kulhanek2021visual} to socially compliant navigation in dynamic crowds \cite{Jin2020SocialSafety}, \cite{Everett2021Collision}, and further to real-world settings with more stringent safety and feasibility requirements \cite{Patel2021DWARL}, \cite{Xie2023DRLVO}.

\begin{figure}[t]
	\centering
	\includegraphics[width=0.98\linewidth]{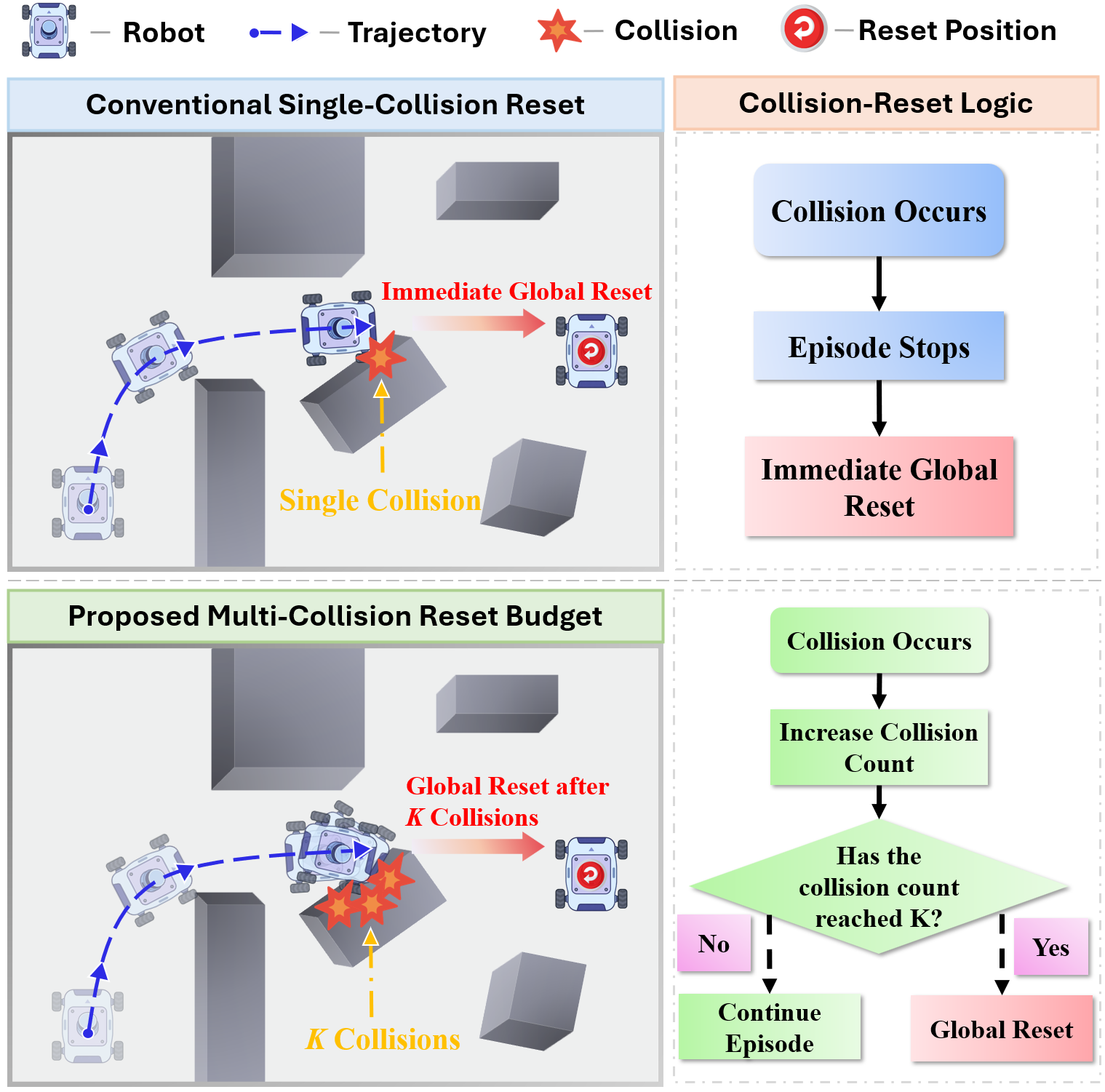}
    \caption{\textbf{Overview of Single-Collision Reset and Multi-Collision Budget.} The conventional protocol immediately triggers a collision-induced global reset after the first collision. In contrast, the proposed multi-collision budget treats collisions before the budget is exhausted as local terminations, allows continued interaction in the same scene, and triggers the collision-induced global reset when the collision count reaches $K$.}
	\label{fig:overview}
\end{figure}

Despite these advances, training efficiency during the early stage of learning remains a known bottleneck. Prior work has explored several directions to address this, including refinement of the observation representation \cite{Zhang2022IPAPRec} and the introduction of guided or assisted exploration \cite{Xie2018TrainingWheels}, \cite{Jang2022Hindsight}. While these efforts report measurable gains, they all share the same underlying assumption that every collision immediately terminates an episode, and none of them questions this default convention itself. More broadly, RL research has examined episode management as a non-trivial component of the training pipeline, ranging from learned reset policies \cite{Eysenbach2018LeaveNoTrace} to reset-free training \cite{Gupta2021ResetFreeMTL}. However, these studies have been developed primarily in the context of general RL or manipulation tasks, and the specific role of collision-triggered resets in DRL-based navigation has received little attention. 
The gap is important because collision handling serves different roles in deployment and training. In deployment, a collision naturally indicates task failure. During training, however, a collision may reflect a local maneuvering error rather than a reason to immediately abandon the scene. Continuing the rollout can provide additional samples around difficult obstacle configurations, whereas the standard immediate-reset convention removes these samples precisely when collisions are most frequent, slowing early-stage policy improvement.

Rethinking the role of collisions in navigation training, we argue that decoupling local collision termination from global environment resets can improve early-stage learning efficiency. In the proposed framework, each collision is still explicitly penalized and treated as a local termination signal, whereas the environment is reset only when the cumulative number of collisions within an episode reaches a predefined collision budget. This mechanism allows the agent to collect additional experience within difficult obstacle configurations that would otherwise be discarded after the first contact. As an optional practical refinement, a pose-change-based filtering strategy is further examined to mitigate the effect of highly redundant collision samples on learning. The main contributions of this letter are summarized as follows:

1) \textbf{The common single-collision reset convention in DRL-based navigation is identified as an overlooked training-design assumption.} By relaxing the coupling between collision events and global environment resets, this work shows that early-stage training efficiency can be improved. To our knowledge, this is the first systematic study of collision-triggered reset handling in DRL-based navigation.

2) A multi-collision reset budget framework is proposed to decouple local collision termination from global environment reset. The framework is paired with a collision-aware transition construction scheme, in which collision transitions are marked as terminal and immediate non-collision bridge transitions are omitted. 

3) Experiments are conducted on multiple simulation platforms and real-world robot platforms with different morphologies. Simulation results show faster early-stage success-rate growth and improved fixed-budget navigation efficiency, while real-world deployments validate the deployability of the learned policies in cluttered environments.




\section{Related Work}

\subsection{DRL-Based Robot Navigation and Efficient Training}

Deep reinforcement learning is widely adopted for robot navigation, where a policy maps local sensory observations directly to low-level motion commands without explicit global planning. Tai \textit{et al.} \cite{Tai2017Virtual} first established a virtual-to-real pipeline from sparse 2D laser readings. Building on this formulation, Patel \textit{et al.} \cite{Patel2021DWARL} coupled the dynamic window approach with DRL to enforce dynamically feasible control among mobile obstacles. Beyond task extension, another group of studies targets training efficiency and generalization. Tobin \textit{et al.} \cite{Tobin2017DomainRandomization} showed that domain randomization enables zero-shot transfer, and a data-efficient framework \cite{Bharadhwaj2019DataEfficient} combines limited expert data with on-policy interaction to reduce sim-to-real cost. LiDAR scans are preprocessed in \cite{Zhang2022IPAPRec} through an adaptively parametric reciprocal function to obtain a better-scaled representation, whereas a hindsight intermediate-target scheme \cite{Jang2022Hindsight} relabels waypoints to densify sparse-reward signals. Safety properties are refined online in \cite{Marzari2023CROP}; hierarchical controllers with novelty-aware exploration \cite{Hu2025LocalOptima} alleviate local-optima trapping; and scenario augmentation \cite{Wang2025ScenarioAug} diversifies the training distribution for better generalization. More recent work has also explored how to reuse informative training experience, using reset curricula that revisit collision-related states \cite{Chen2026SEANav} or predictive trajectory reuse to improve value learning \cite{romero2025actor}.

\subsection{Episode Management in Reinforcement Learning}

Episode boundary handling governs which transitions are stored, how bootstrap targets are computed, and when the agent is relocated. This aspect remains far less studied than policy design or reward shaping. Pardo \textit{et al.} \cite{Pardo2018TimeLimits} distinguished genuine terminations from training-induced truncations and showed that conflating the two biases value estimation, indicating that value-learning boundaries should be decoupled from artificial episode resets. A complementary line questions the reliance on externally triggered resets. Eysenbach \textit{et al.} \cite{Eysenbach2018LeaveNoTrace} jointly learned a task policy and a reset policy so that the agent returns to safe initial states autonomously, whereas Recovery RL \cite{Thananjeyan2021RecoveryRL} decoupled task execution from recovery through learned recovery zones activated before imminent constraint violations. Removing fixed re-initialization altogether, a reset-free trial-and-error paradigm \cite{Chatzilygeroudis2018ResetFree} enables continual adaptation, and multi-task coupling \cite{Gupta2021ResetFreeMTL} arranges tasks so that the completion of one implicitly resets another. On the scheduling side, \cite{Bharthulwar2025Staggered} showed that synchronized resets across parallel environments induce harmful nonstationarity, whereas staggered schedules restore temporal diversity. How collisions should terminate episodes in DRL-based  navigation, however, remains systematically unexamined.

\section{Preliminaries}



\subsection{State Representation and Reward Design}

The policy is trained with Soft Actor-Critic (SAC)~\cite{haarnoja2018soft}, an off-policy actor-critic algorithm that operates on transitions stored in a replay buffer. The state $s_t$ encodes real-time 2D LiDAR readings, the robot velocity, and the goal position expressed in the body frame, and is given by:
\begin{equation}
  s_t = \big[\hat{l}_1, \ldots, \hat{l}_m, \; d^g_t, \; \varphi^g_t, \; v_t, \; \omega_t \big],
\end{equation}
where $\hat{l}_1, \ldots, \hat{l}_m$ are processed LiDAR readings, $d^g_t$ and $\varphi^g_t$ are the relative distance and angle to the goal, and $v_t$ and $\omega_t$ are the current linear and angular velocities. Each LiDAR reading is transformed by an adaptively parametric reciprocal function $\hat{l}_j = 1/(\bar{l}_j - \beta)$ following~\cite{Zhang2022IPAPRec}, where $\beta$ is a trainable parameter updated jointly with the policy network. The reward function is given by:
\begin{equation}
  r_t =
  \begin{cases}
    r_{\text{success}}, & \text{if the goal is reached,} \\
    r_{\text{collision}}, & \text{if a collision occurs,} \\
    c_1 (d^g_t - d^g_{t+1}), & \text{otherwise,}
  \end{cases}
\end{equation}
where $r_{\text{success}}=10$ is a positive reward for goal arrival, $r_{\text{collision}}=-10$ is a negative collision penalty, $c_1$ is a scaling constant, and the third term is a dense shaping signal encouraging progress toward the goal.

\subsection{Local Termination and Global Reset}
\label{sec:local_global}

In DRL-based navigation, each episode may end with one of three events: the robot reaches the goal (success), collides with an obstacle (collision), or exceeds the maximum time step limit $T_{\max}$ (timeout). 
To clarify how these events are handled in the proposed framework, we distinguish between two notions: local termination and global reset. Local termination refers to a collision that is treated as a local failure signal: the agent receives a collision penalty, but the episode is not necessarily terminated and the robot is allowed to continue interacting with the same scene. Global reset, in contrast, refers to the reinitialization of the environment, which re-randomizes the robot starting pose and goal position to begin a new episode. In the proposed framework, global reset is triggered by success, timeout, or the accumulation of collisions reaching the collision budget, whereas local termination corresponds to the collisions that occur before this budget is exhausted. Conventional DRL navigation frameworks couple every collision with an immediate global reset, which discards the learning opportunity within the same scene; this conflation of local failure supervision with global environment resets motivates the method presented in the following section.

\section{Approach}
\label{sec:method}

\subsection{Multi-Collision Reset Budget}
\label{sec:budget}

The core idea of the proposed framework is to allow the agent to sustain multiple collisions within a single episode before a global reset is triggered. To this end, we introduce a collision counter $c$ that records the cumulative number of collisions within the current episode, alongside a collision budget $K$ that specifies the maximum number of local failures permitted before the environment is reinitialized.

At the start of an episode, the collision counter is initialized as $c=0$. After each collision, the counter is updated as $c \leftarrow c+1$. 
For value learning, all stored collision transitions are assigned a terminal flag $d_t=1$, regardless of whether the collision triggers a global reset. 
If $c<K$, this terminal event remains local: the environment is not reinitialized, the robot is not relocated to a newly sampled pose, the goal remains unchanged, and the obstacle layout is preserved. At the next control step, the policy receives the post-collision observation and continues the navigation task in the same scene. No external recovery controller or manual unsticking operation is applied before the collision budget is exhausted. 
The finite collision budget bounds this retry process, preventing the agent from spending excessive interaction time in unrecoverable or highly redundant contact states while still allowing limited post-collision exploration. If $c \ge K$, a global reset is triggered and a new randomized scenario is sampled. Success and timeout keep their standard semantics: reaching the goal or exceeding $T_\text{max}$ always triggers a global reset, independent of the current collision count.

While this mechanism is simple in form, two practical issues need to be handled. Since one episode can contain several collisions, replay-buffer entries must keep value targets consistent across local termination boundaries. Otherwise, transitions after a collision may create incorrect bootstrap targets and blur credit assignment near obstacles. Repeated contacts in the same scene can also add nearly identical failure samples, especially when the robot is stuck and keeps moving with similar headings. Such samples can skew the replay-buffer distribution toward redundant collision experience.

\begin{algorithm}[t]
\caption{Training procedure of MCB.}
\label{alg:mcb}
\begin{algorithmic}[1]
\State \textbf{Input:} budget $K$, horizon $T_{\max}$.
\State \textbf{Initialize:} $\pi_{\theta}$, $Q_{\phi_1}$, $Q_{\phi_2}$, replay buffer $\mathcal{B}$.
\For{$\text{episode}=1,2,\ldots$}
  \State Reset environment and obtain initial observation $x_0$.
  \State Set collision count $c\gets0$ and $b\gets\text{False}$.
  \LineComment{// $b$: whether the previous event was a collision.}
  \For{$t=1,2,\ldots,T_{\max}$}
    \State Sample $a_t\sim\pi_{\theta}(\cdot|s_t)$; receive $(s_{t+1},r_t,e_t)$.
    \State $d_t\gets\mathbb{I}[e_t\in\{\text{success},\text{collision}\}]$.
    \State $\beta_t^{\mathrm{br}}\gets\mathbb{I}[b\wedge(e_t\ne\text{collision})]$.
    \LineComment{// $\beta_t^{\mathrm{br}}$: non-collision bridge flag.}
    \State \textbf{if} $\beta_t^{\mathrm{br}}=0$ \textbf{then} add $(s_t,a_t,r_t,s_{t+1},d_t)$ to $\mathcal{B}$.
    \State $c\gets c+\mathbb{I}[e_t=\text{collision}]$.
    \LineComment{// $c<K$: local termination; $c\ge K$: global reset.}
    \State Sample mini-batch $\mathcal{M}\sim\mathcal{B}$ and update $\theta,\phi_1,\phi_2$.
    \State \textbf{if} $e_t=\text{success}$ \textbf{or} $c\ge K$ \textbf{or} $t=T_{\text{max}}$ \textbf{then break}.
    \State $b\gets\mathbb{I}[e_t=\text{collision}]$.
  \EndFor
\EndFor
\State \Return $\pi_{\theta}$.
\end{algorithmic}
\end{algorithm}

\subsection{Collision-Aware Transition Construction}
\label{sec:transition}
Decoupling local termination from global reset requires careful  treatment of the transitions stored in the replay buffer, ensuring that the allowance of post-collision exploration does not introduce conflicting learning signals.  
Every collision transition is marked with a terminal flag $d_t = 1$, whether or not it triggers a global reset. With this flag, the Bellman backup target at the collision step reduces to the immediate reward $r_t$, because the terminal indicator drops the bootstrapped value of the next state. The collision penalty is therefore absorbed at the step where the failure happens, rather than diluted by rewards collected after the agent resumes navigation. This keeps credit assignment clean: the critic attributes the collision cost to the state-action pairs near the obstacle that caused the failure, without mixing them with returns earned after resumption.

The terminal flag blocks value propagation across a collision boundary, but it does not specify how the first transition after a non-resetting collision should be handled in the replay buffer. Consider a collision at step $t$, before the collision count reaches the reset budget. The collision transition is stored as $(s_t,a_t,r_t,s_{t+1},d_t=1)$, so its Bellman target consists only of the immediate collision penalty and does not bootstrap from $s_{t+1}$. Because this collision does not trigger a global reset, the rollout continues from the simulator-resolved post-collision state $s_{t+1}$.
The transition immediately after this collision is handled according to its outcome. If the next action also causes a collision, the transition is retained and marked as terminal. This sample represents another unsafe action taken from the current contact state and should therefore contribute its own collision penalty. If the next action does not cause a collision, the transition is instead treated as a bridge transition. Adding it to the replay buffer as a regular sample would make the simulator-resolved post-collision state $s_{t+1}$ an ordinary bootstrapped decision state, allowing value estimates to propagate from later non-collision states. 
This would make post-collision continuations part of the training distribution, although such continuations are outside the evaluation protocol, where any collision terminates the navigation trial. For this reason, only the immediate non-collision bridge transition after a local collision is omitted from the replay buffer. This affects replay construction only, the rollout continues in the same scene. The complete training procedure is summarized in Algorithm~\ref{alg:mcb}.
\subsection{Pose-Change-Based Collision Filtering}

A lightweight filtering mechanism is added as an optional refinement to the multi-collision reset budget to control the collision transitions stored in the replay buffer. This mechanism addresses a side effect that can occur when $K > 1$. When multiple local terminations are allowed within one episode, the agent may produce several consecutive collisions with nearly unchanged headings, especially when it is stuck against an obstacle and repeatedly executes similar actions. These transitions often correspond to the same failed behavior rather than distinct navigation attempts, and storing many of them may reduce the diversity of obstacle-avoidance experience available for learning.
To reduce this redundancy, collision transitions are filtered according to the change in robot heading. Let $\theta_t$ denote the robot heading at time step $t$. For each collision transition, the angular displacement from the previous collision is computed as
\begin{equation}
\Delta \theta_t = \left| \mathrm{atan2}\left( \sin(\theta_t - \theta_{t-1}),\ 
\cos(\theta_t - \theta_{t-1}) \right) \right|,
\end{equation}
where the $\mathrm{atan2}$ form accounts for angle wrapping. A collision transition is stored in the replay buffer only when $\Delta \theta_t \geq \tau_\theta$, where $\tau_\theta$ is a predefined angular threshold. Transitions with $\Delta \theta_t < \tau_\theta$ are discarded as redundant. This criterion preserves collisions that occur from sufficiently different approach directions while suppressing repeated contacts under similar orientations.

\begin{figure}[t]
	\centering
  \subfloat[Training Env\_1]{
  \centering\includegraphics[width=0.311\linewidth]{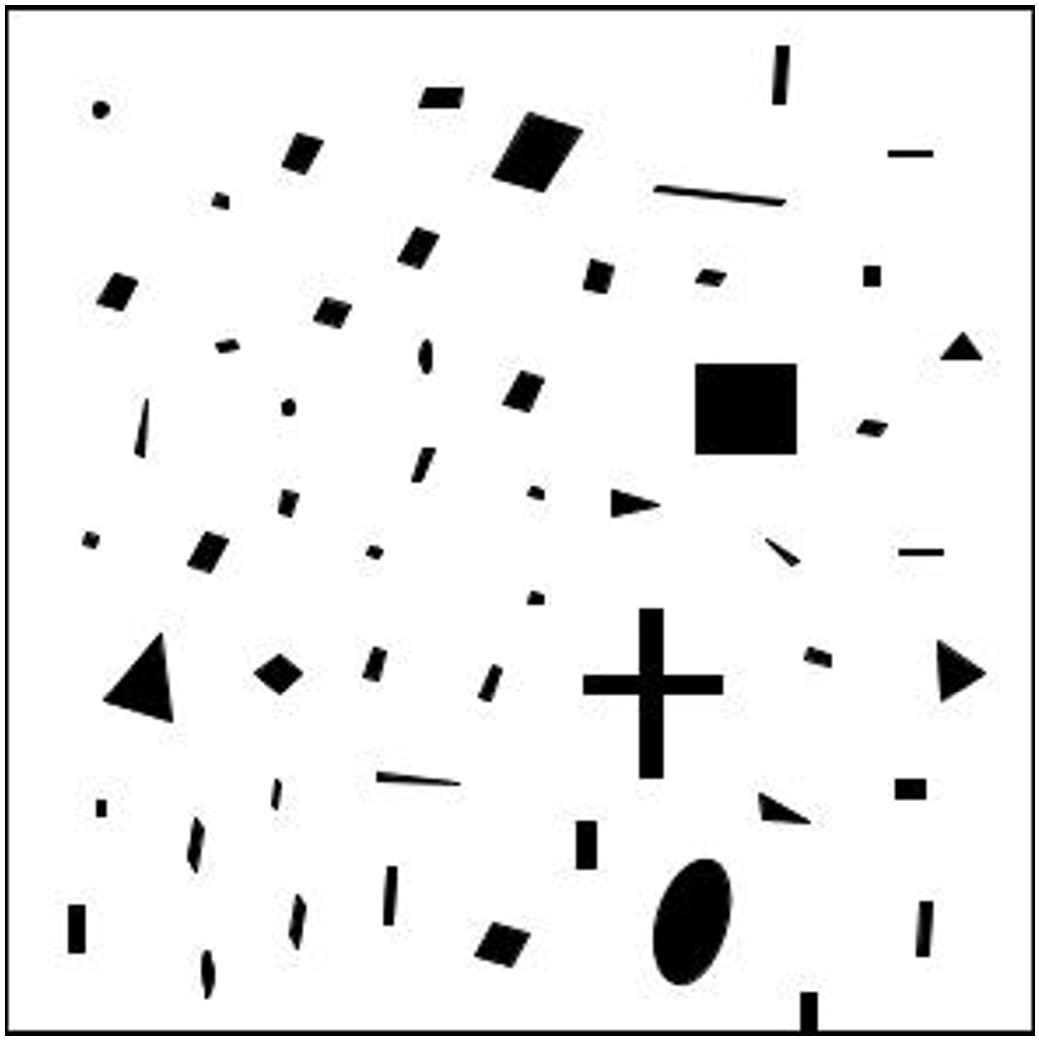}\label{trainingenv}}
  \subfloat[Testing Env\_2]{
  \centering\includegraphics[width=0.308\linewidth]{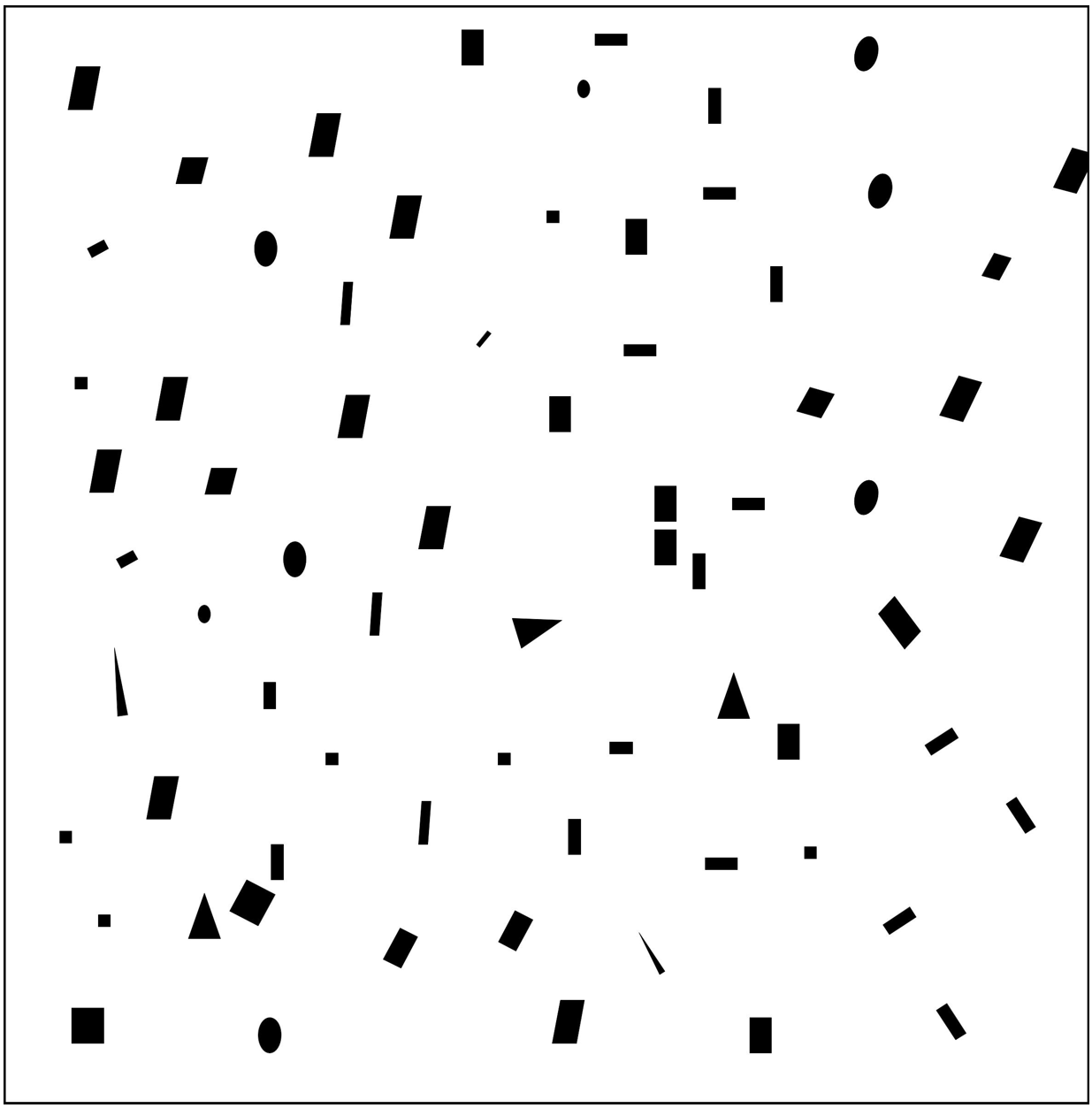}\label{testenv}}
  \subfloat[Isaac Lab Scene]{
  \centering\includegraphics[width=0.31\linewidth]{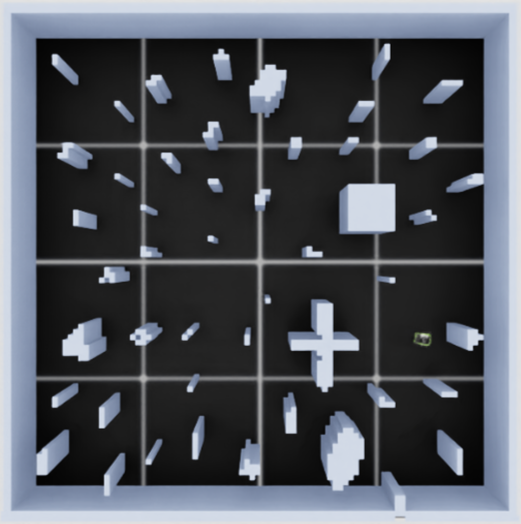}\label{Isaac Lab Scene}}
	\caption{Training environment Env\_1 and an unseen test environment Env\_2 in Stage, with a corresponding scene example in Isaac Lab.}
	\label{real_test_env}
\end{figure}

\section{Simulation Results and Analysis}

\subsection{Experimental Setup}

\subsubsection{Implementation Details}
All experiments are conducted on a single PC equipped with an NVIDIA RTX 4090 GPU. Two simulation platforms are employed: the Stage simulator~\cite{stagesimulator}, and the 
Isaac Lab platform~\cite{isaac}. 
The training environment and the unseen testing environment are illustrated in Fig.~\ref{trainingenv} and Fig.~\ref{testenv}, respectively. The policy network and the critic networks are implemented as four-layer multilayer perceptrons with 100 hidden units per layer. The action space consists of the linear velocity bounded within $[0,\,0.5]$\,m/s and the angular velocity bounded within $[-\pi/2,\,\pi/2]$\,rad/s. In Stage, the maximum episode length is set to 200 steps, and each method is trained for 50{,}000 steps, with a checkpoint saved every 2{,}500 steps for evaluation. In Isaac Lab, training is organized into epochs, where one epoch corresponds to one rollout-update cycle over 512 parallel environments with 16 interaction steps collected per environment. Each method is trained for 3{,}000 epochs, and evaluation is performed every 150 epochs. The maximum episode length is kept at 200 steps on both platforms. 
To reduce seed-specific effects, each method is trained with 10 random seeds, and all reported curves and tables are averaged over those runs. Each checkpoint is evaluated on the same set of 50 point-to-point navigation tasks. 

\begin{table}[t]
\centering
\caption{Summary of the compared methods.}
\label{tab:method_naming}
\setlength{\tabcolsep}{12pt}
\renewcommand{\arraystretch}{1.18}
\begin{tabular}{@{}c c@{}}
\toprule
Method & Note \\
\midrule
SCR 
& \textbf{S}ingle-\textbf{C}ollision \textbf{R}eset baseline. \\
\addlinespace[2pt]
MCB-K$K$ 
& \textbf{M}ulti-\textbf{C}ollision \textbf{B}udget with budget $K$. \\
\addlinespace[2pt]
MCB-K$K$-PF$\theta$
& \begin{tabular}[c]{@{}c@{}}
MCB-K$K$ with additional \textbf{P}ose-change \\
\textbf{F}iltering at angular threshold $\tau_{\theta}$.
\end{tabular} \\
\bottomrule
\end{tabular}
\end{table}

\begin{table}[t]
\centering
\caption{Navigation performance at the final training checkpoint in the Stage environments.}
\label{mainresult}
\setlength{\tabcolsep}{3.8pt}
\renewcommand{\arraystretch}{1.12}
\begin{tabular}{@{}c*{8}{c}@{}}
\toprule
\multirow{2}{*}{Method} 
& \multicolumn{4}{c}{Training Env\_1} 
& \multicolumn{4}{c}{Testing Env\_2} \\
\cmidrule(lr){2-5} \cmidrule(lr){6-9}
& SR$\uparrow$ & AV$\uparrow$ & AEL$\downarrow$ & ANS$\uparrow$
& SR$\uparrow$ & AV$\uparrow$ & AEL$\downarrow$ & ANS$\uparrow$ \\
\midrule
\addlinespace[2pt]
SCR        
& 0.93 & 0.39 & 87.45 & 0.71 
& 0.74 & 0.34 & 82.44 & 0.54 \\
MCB-K2     
& \textbf{0.96} & 0.40 & \textbf{57.40} & \textbf{0.81} 
& 0.80 & 0.37 & \textbf{51.39} & \textbf{0.67} \\
MCB-K2-PF3 
& \textbf{0.96} & \textbf{0.43} & 77.48 & 0.77 
& \textbf{0.81} & \textbf{0.40} & 66.78 & 0.65 \\
\bottomrule
\end{tabular}
\end{table}

\subsubsection{Evaluation Metrics}

Navigation performance is characterized using four metrics, organized according to the aspect of behavior they measure. During evaluation, the relaxed collision handling used during training is disabled; any collision terminates the episode and is treated as a task failure. The first group captures task-level outcomes. The success rate (SR) is the fraction of evaluation episodes in which the robot reaches the goal without collision or timeout. The average navigation score (ANS) combines task completion and time cost into a single scalar for comparing methods, and is defined as

\begin{equation}
S =
\begin{cases}
1 - \dfrac{2T_{s}}{T_{\text{max}}}, & \text{if success}, \\[4pt]
-1, & \text{otherwise},
\end{cases}
\label{eq:nav_score}
\end{equation}
where $T_{s}$ denotes the number of navigation steps consumed in the current episode and $T_{\text{max}}$ denotes the maximum episode length. The second group reports motion-level behavior. The average velocity (AV) is the mean linear velocity of the robot across all evaluation steps, and the average episode length (AEL) is the mean number of steps elapsed before an episode terminates.

\subsection{Overall Performance and Learning Efficiency in Stage}
\label{subsec:overall}

Following the naming conventions in Table~\ref{tab:method_naming}, three representative methods are evaluated: SCR, MCB-K2 as the small-budget MCB instance, and MCB-K2-PF3 as the pose-change-filtered variant with $\tau_{\theta}=3^{\circ}$, chosen for its best observed trade-off between robustness across collision budgets and navigation performance.

\begin{figure}[t]
	\centering
	\includegraphics[width=0.98\linewidth]{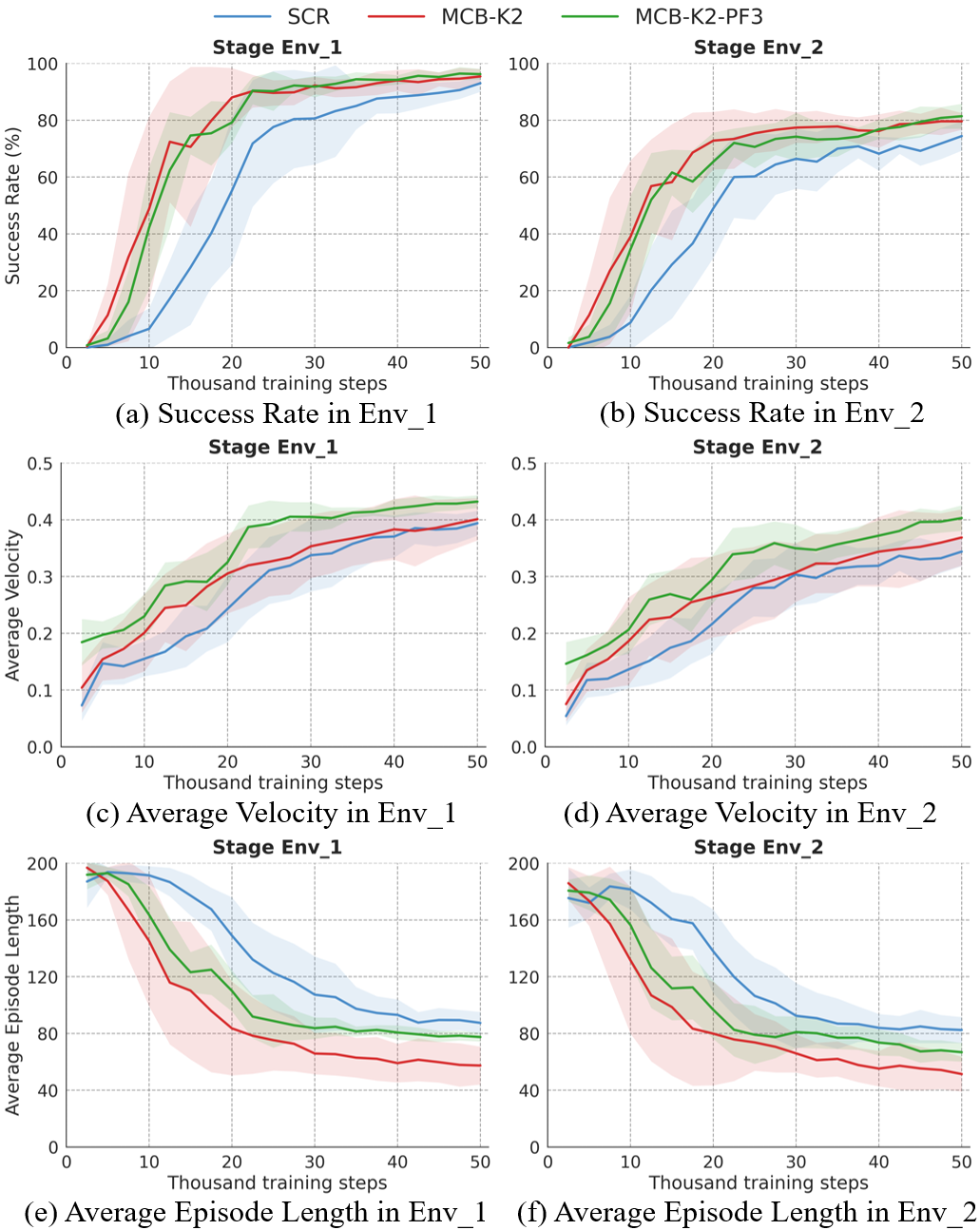}
\caption{Learning curves and navigation efficiency metrics of the compared methods on Env\_1 and Env\_2 in Stage. Solid lines and shaded regions represent the mean and standard deviation over ten random seeds.}
	\label{mainmetric}
\end{figure}

\subsubsection{Fixed-Budget Navigation Performance}

Table~\ref{mainresult} reports higher performance for both MCB variants than SCR on all metrics in both environments. The three methods use the same network architecture, reward design, and training budget, and differ only in how collision events are coupled to environment reset. The results therefore isolate collision-reset coupling as a factor that affects the final navigation policy, rather than an effect of model capacity or training budget. MCB-K2 improves the success rate in both environments and produces the most time-efficient behavior, with the shortest average episode length and the highest ANS in the unseen Env\_2. MCB-K2-PF3 attains a comparable success rate and a higher average velocity, suggesting that the pose-change filter changes the learned motion profile without reducing task completion. This filter is therefore useful when higher travel speed is desired.

\subsubsection{Early-Stage Learning Efficiency}

\begin{figure}[t]
	\centering
	\includegraphics[width=0.98\linewidth]{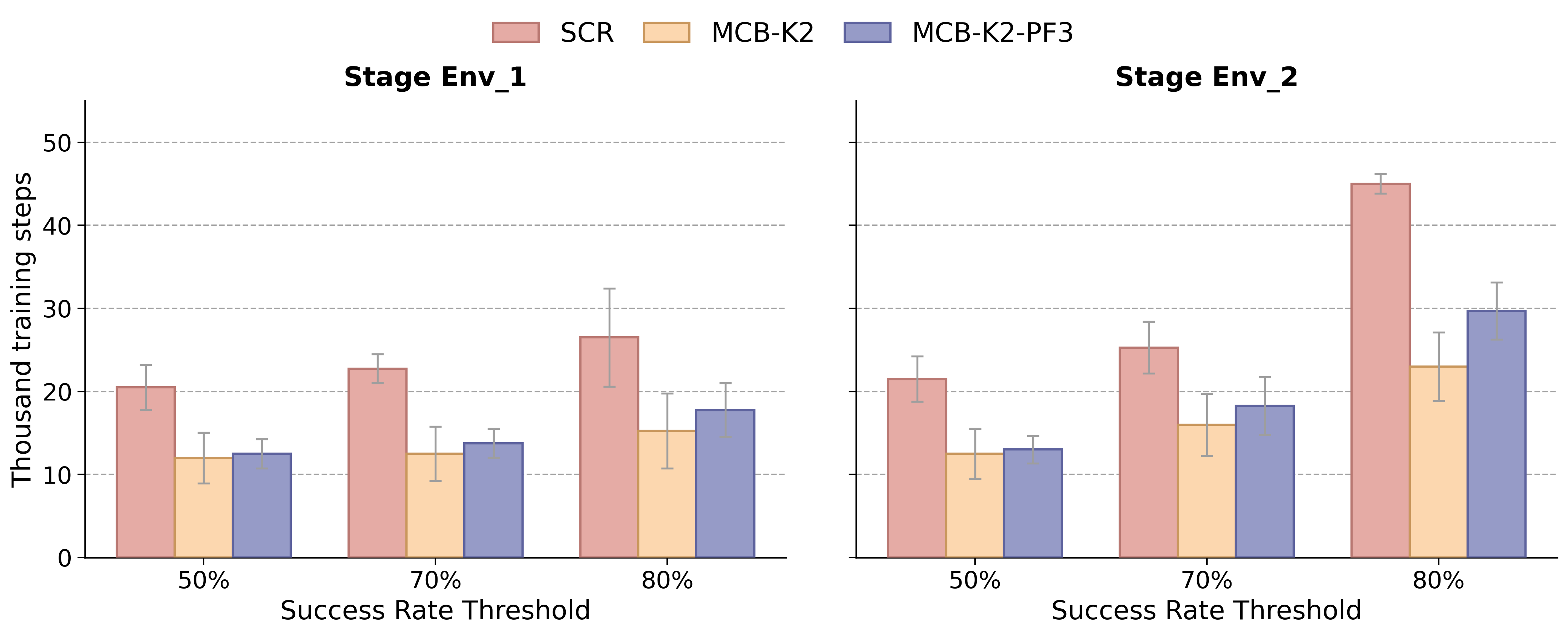}
	\caption{Training steps at which the evaluated success rate first reaches 50\%, 70\%, and 80\% thresholds in Env\_1 and Env\_2. Bars and error bars indicate the mean and standard deviation over ten random seeds, respectively.}
	\label{threshold_steps}
\end{figure}

The advantage of MCB over SCR is most pronounced in the early stage of training. In Fig.~\ref{mainmetric}, the success rates of MCB-K2 and MCB-K2-PF3 increase rapidly and reach high levels before SCR shows a similar rise. A comparable gap is also observed in average velocity and episode length. Fig.~\ref{threshold_steps} reports the number of training steps required to first reach $50\%$, $70\%$, and $80\%$ success. Across both environments and all three thresholds, MCB-K2 reaches each milestone at least $36.7\%$ earlier than SCR. MCB-K2-PF3 also reduces the required steps in every case, with the smallest reduction being $27.8\%$. These results indicate that separating local collision termination from global reset mainly improves early-stage learning efficiency, as shown by faster success-rate growth and fewer interactions needed to reach the same success thresholds.

\begin{figure}[t]
	\centering
	\includegraphics[width=0.98\linewidth]{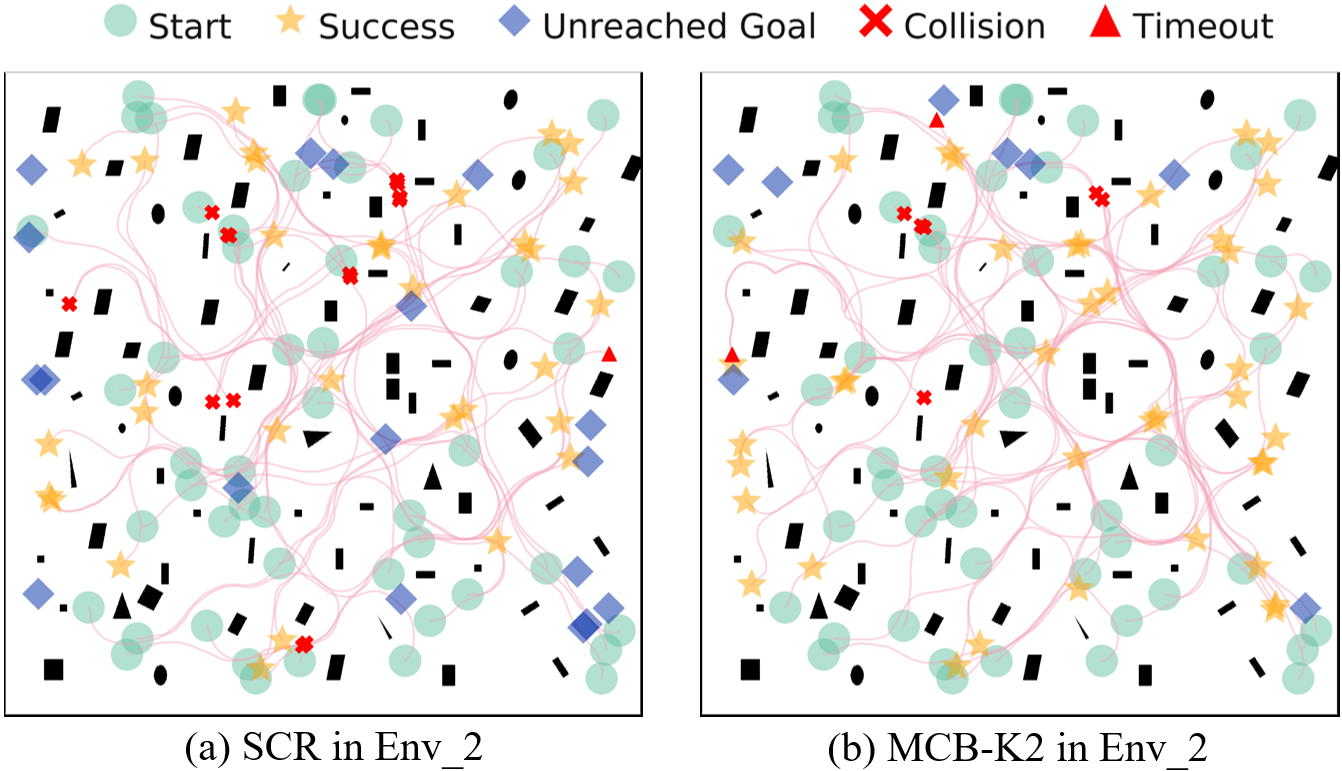}
\caption{Trajectory comparison between SCR and MCB-K2 in Env\_2.}
	\label{trajcompare}
\end{figure}

\subsubsection{Trajectory-Level Behavior}

Fig.~\ref{trajcompare} compares the evaluation trajectories of SCR and MCB-K2 in Env\_2. SCR often fails in narrow passages between obstacles, with collisions concentrated in regions that require accurate local maneuvers. MCB-K2 passes through the same regions with fewer collisions and follows smoother trajectories toward the goal.

\begin{figure}[t]
    \centering
    \includegraphics[width=0.98\linewidth]{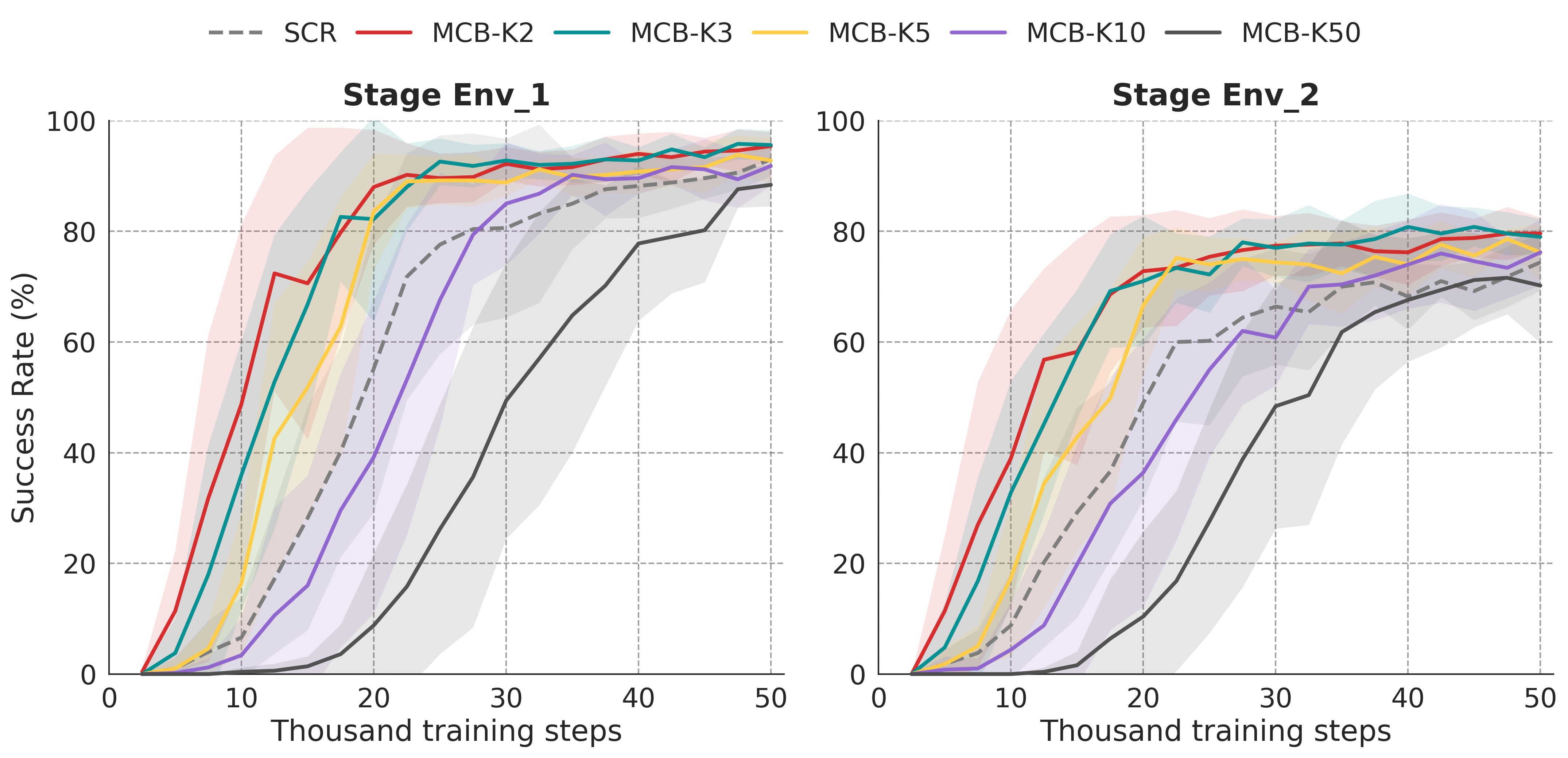}
    \caption{Learning curves of MCB with collision budget $K \in \{2, 3, 5, 10, 50\}$ and the SCR baseline on Env\_1 and Env\_2 in Stage.}
    \label{Kablation}
\end{figure}

\begin{table}[t]
\centering
\caption{Range of final-checkpoint success rates across collision budgets under different pose-change filter settings.}
\label{theta_robustness}
\setlength{\tabcolsep}{9.5pt}
\renewcommand{\arraystretch}{1}
\begin{tabular}{@{}ccc@{}}
\toprule
\multirow{2}{*}{\centering PF setting}
& \multicolumn{2}{c}{SR Range (\%) $\downarrow$} \\
\cmidrule(l{2pt}r{2pt}){2-3}
& Training Env\_1 & Testing Env\_2 \\
\midrule
w/o PF                       & 7.2          & 9.4          \\
$\tau_{\theta}=0.5^{\circ}$  & 1.8          & 2.0          \\
$\tau_{\theta}=1^{\circ}$    & 3.0          & 3.0          \\
$\tau_{\theta}=2^{\circ}$    & \textbf{1.6} & \textbf{1.8} \\
$\tau_{\theta}=3^{\circ}$    & 3.0         & 4.6          \\
$\tau_{\theta}=10^{\circ}$   & 2.4          & 3.6          \\
\bottomrule
\end{tabular}
\end{table}

\subsection{Ablation Study}
\label{subsec:ablation}

The ablation analysis focuses on the two design parameters of the proposed framework: the collision budget $K$ in the multi-collision reset rule and the angular threshold $\tau_{\theta}$ in the optional pose-change-based filter.

\subsubsection{Effect of the Collision Budget}
\label{subsubsec:K_ablation}

The MCB framework is trained with $K \in \{2,3,5,10,50\}$ without the pose-change filter and compared against SCR. Figure~\ref{Kablation} shows a clear dependence on $K$. Small budgets lead to faster early learning and higher success rates in both environments, whereas larger budgets reduce the early-stage advantage and lower final success. At the largest budget, the success rate remains below that of SCR throughout training. This pattern suggests that an overly large per-episode collision budget introduces repeated and highly correlated failure transitions into the replay buffer, which reduces the relative share of informative samples and slows value learning.

\subsubsection{Robustness of the Pose-Change Filter}
\label{subsubsec:PF_ablation}

The influence of the optional filter on sensitivity to $K$ is evaluated using the range of converged success rates over $K \in \{2,3,5,10,50\}$, computed as the difference between the maximum and minimum values. A smaller range indicates weaker dependence on $K$. Table~\ref{theta_robustness} reports this range in both environments for the unfiltered framework and five $\tau_{\theta}$ settings. Without filtering, the success rate changes substantially with $K$, indicating strong dependence on the collision budget. After applying the pose-change filter, the range decreases for all tested $\tau_{\theta}$ values, while the best success rate in each environment changes only slightly across thresholds. The filter therefore reduces sensitivity to the collision budget and is not strongly affected by the exact choice of $\tau_{\theta}$ within the tested range, making it a practical optional refinement with limited tuning cost.

\subsubsection{Replay-Level Collision Statistics}
\label{subsubsec:replay_stats}

Replay composition under the pose-change filter is further examined through collision statistics from MCB-K$K$-PF3 with $\tau_{\theta}=3^{\circ}$, the setting used in the main comparison. Table~\ref{tab:replay_collision_stats} shows that increasing $K$ substantially increases both the number of collision candidates and their candidate ratio. The PF-filtered ratio also increases, keeping the stored collision ratio below $1\%$ even when $K=50$. These results show that large budgets generate many additional collision candidates, most of which are removed by the pose-change filter as redundant local-failure samples. This replay-level evidence is consistent with the preceding ablations: small budgets limit excessive collision sampling, while pose filtering mitigates replay-buffer imbalance when larger budgets are used.

\begin{figure}[t]
    \centering
    \includegraphics[width=0.98\linewidth]{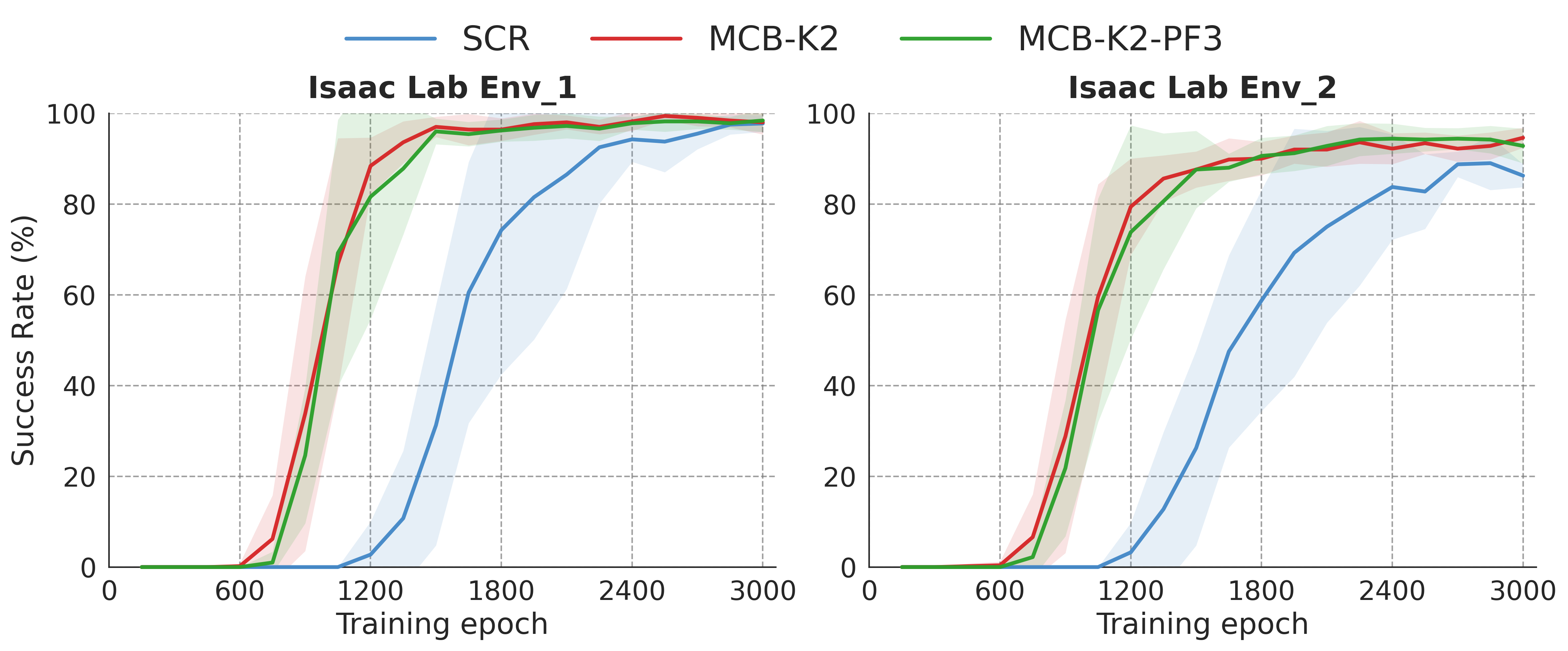}
    \caption{Learning curves of compared methods in Isaac Lab.}
    \label{isaaclabsuccurve}
\end{figure}

\begin{table}[t]
\centering
\caption{Collision Statistics under MCB-K$K$-PF3 with $\tau_{\theta}=3^{\circ}$.}
\label{tab:replay_collision_stats}
\footnotesize
\setlength{\tabcolsep}{5pt}
\renewcommand{\arraystretch}{1.08}
\begin{tabular}{@{}ccccc@{}}
\toprule
\shortstack[c]{Collision\\Budget} 
& \shortstack[c]{Collision\\Candidates} 
& \shortstack[c]{Candidate\\Ratio (\%)} 
& \shortstack[c]{PF-Filtered\\Ratio (\%)} 
& \shortstack[c]{Stored Collision\\Ratio (\%)} \\
\midrule
2  & 231  & 0.44 & 37.02 & 0.28 \\
3  & 346  & 0.66 & 51.48 & 0.32 \\
5  & 602  & 1.15 & 65.95 & 0.39 \\
10 & 1143 & 2.16 & 79.21 & 0.46 \\
50 & 5587 & 9.81 & 93.84 & 0.67 \\
\bottomrule
\end{tabular}
\end{table}

\subsection{Cross-Platform Validation in Isaac Lab}

To verify that the observed improvements are not specific to the Stage simulator and to examine how the proposed framework scales under a modern GPU-accelerated training paradigm, we further validated all three methods on the Isaac Lab platform. As shown in Fig.~\ref{isaaclabsuccurve}, the same trend is reproduced, and the advantage of our method becomes even more pronounced in this parallel training setting. 
Both MCB-K2 and MCB-K2-PF3 reach target success-rate levels earlier than SCR, confirming that the early-stage learning-efficiency gain is reproduced under GPU-accelerated parallel training.

\section{Real-World Experiments}

\begin{figure*}[t]
    \centering
    \includegraphics[width=\linewidth]{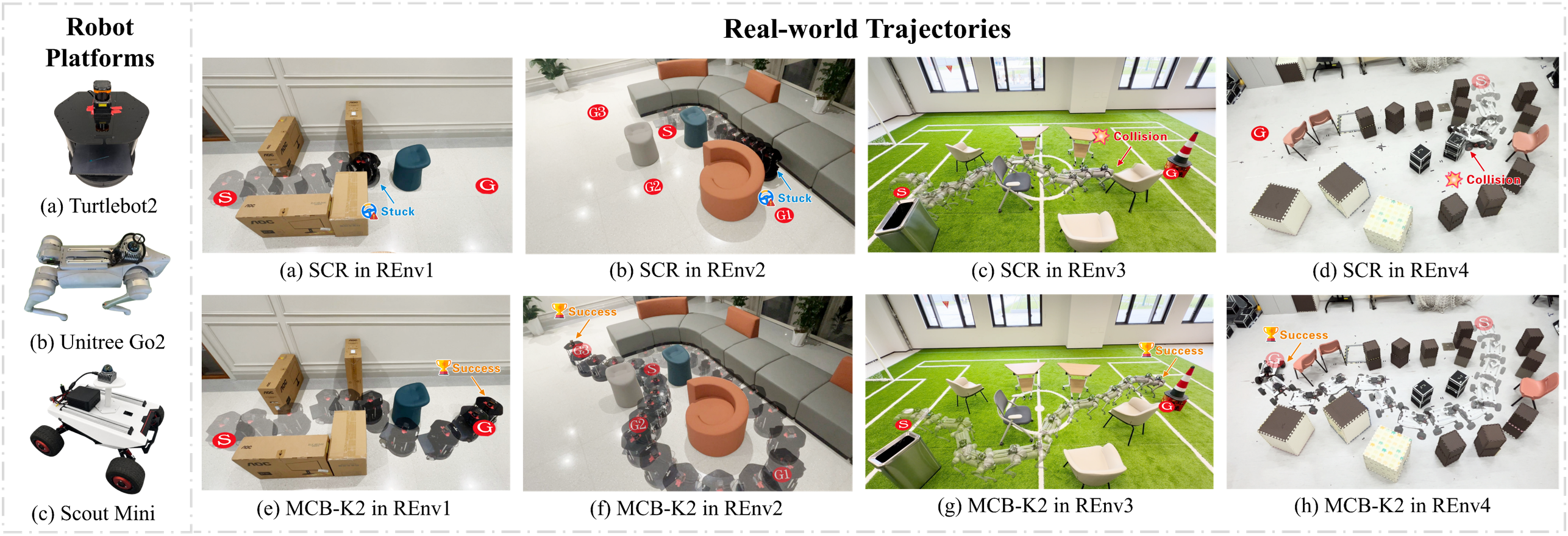}
    \caption{Robot platforms and real-world trajectories of robots trained with SCR and MCB-K2 across four test scenarios. The start point is marked as $S$, and the target point is marked as $G$.}
\label{realtestenv}

\end{figure*}


\subsection{Hardware Setup}

As shown in Fig.~\ref{realtestenv}, the real-world experiments were conducted on three physical robot platforms: a TurtleBot2 wheeled robot, an Agilex Scout Mini wheeled robot, and a Unitree Go2 legged robot. The TurtleBot2 was equipped with a Hokuyo UTM-30LX 2D LiDAR and an onboard PC with an Intel i9-12900H CPU. The Agilex Scout Mini was equipped with a Livox MID-360 3D LiDAR and an Aoostar GEM12 Mini PC as the onboard computer. For these two wheeled platforms, the testing environments were mapped using ROS GMapping~\cite{Gmapping}, and the robot pose was estimated using ROS AMCL~\cite{AMCL}. The map was used only to obtain the target position in the robot frame and was not used for motion planning. The Unitree Go2 was equipped with a Livox MID-360 3D LiDAR and an NVIDIA Jetson Orin NX onboard computer. For this legged platform, target localization was provided by a companion UWB module placed at the target location during evaluation.

\subsection{Real-World Scenarios and Task Design}

The robots were evaluated in four real-world scenarios, REnv1, REnv2, REnv3 and REnv4. REnv1 used intentionally arranged obstacles forming a very narrow passage, simulating dense indoor clutter and testing navigation through highly constrained spaces. REnv2, REnv3 and REnv4 were naturally cluttered offices with moderate congestion, used to assess navigation in realistic workplaces. In all scenarios, the robot started at $S$ and navigated point-to-point to $G$. REnv2 included multiple sequential goals, denoted $G_i$; after each arrival, the robot briefly paused to indicate success before continuing to the next goal.

\subsection{Experimental Results Analysis}

\begin{figure}[t]
	\centering
  \subfloat[Dynamic Scenario in REnv2]{
  \centering\includegraphics[width=0.485\linewidth]{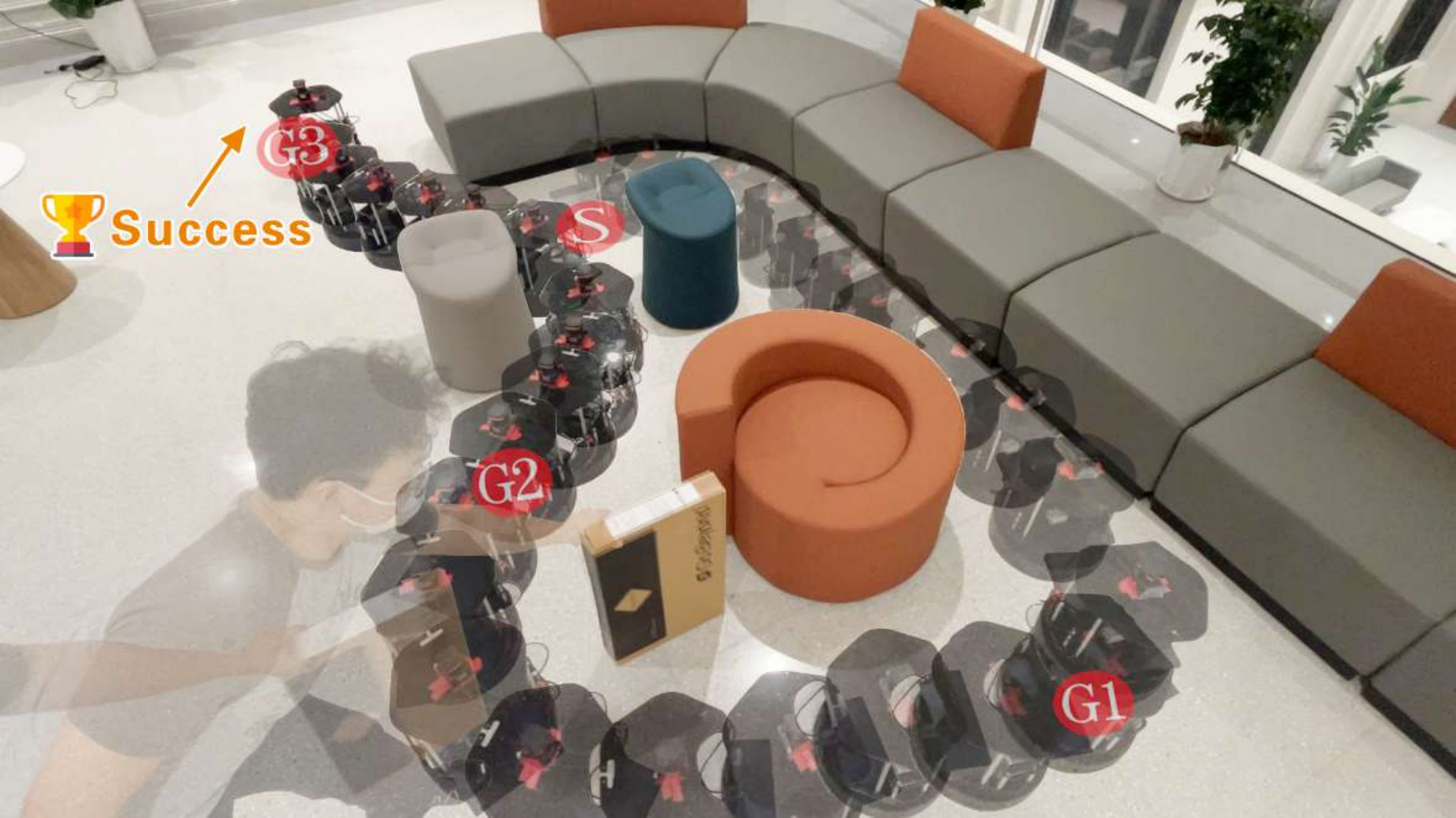}\label{turtlebotdyn}}
  \subfloat[Dynamic Scenario in REnv3]{
  \centering\includegraphics[width=0.488\linewidth]{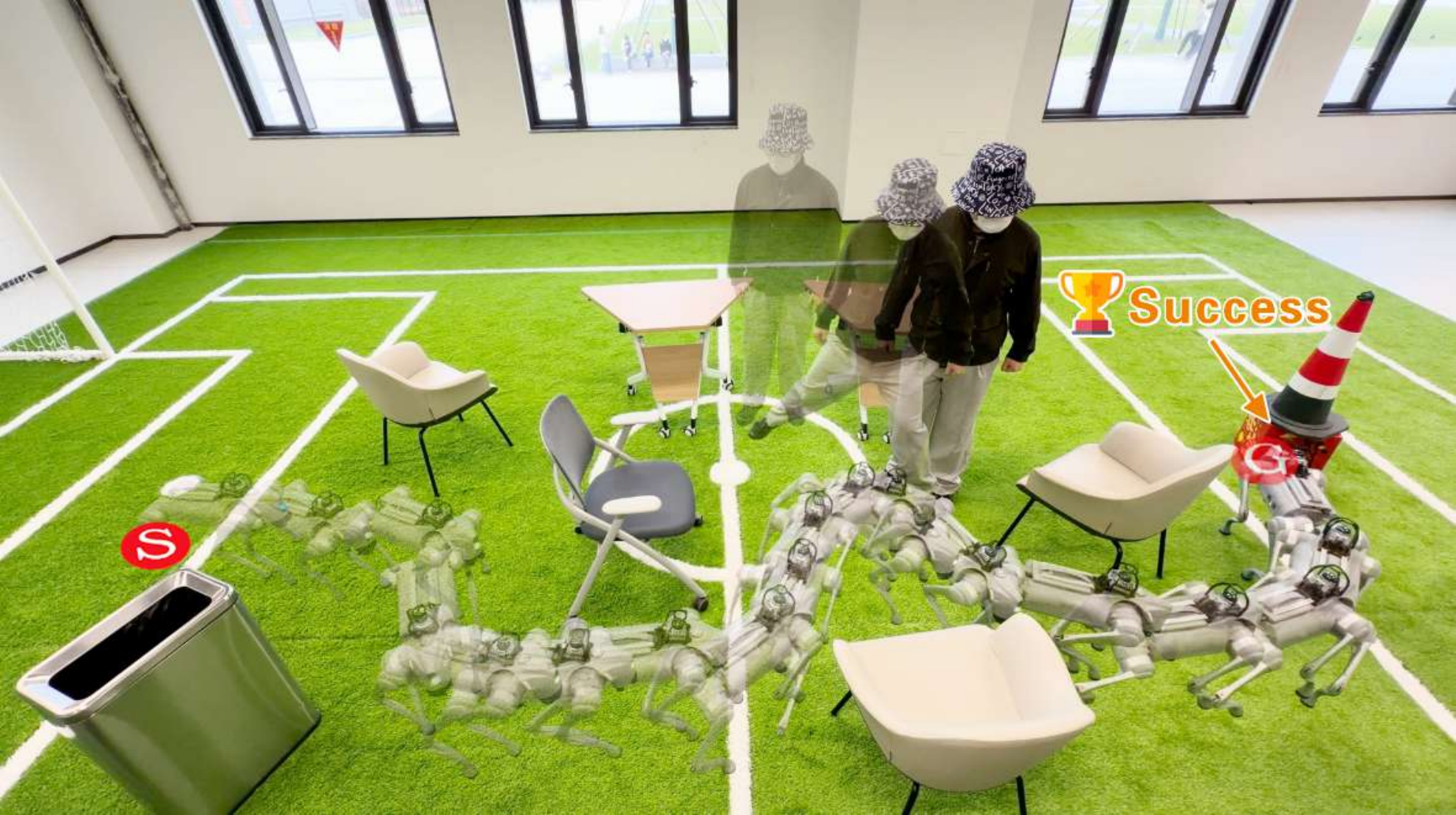}\label{dogdyn}}
	\caption{Dynamic Real-World Navigation Results of MCB-K2.}
	\label{dynamicenv}
\end{figure}

The real-world testing results are shown in Fig.~\ref{realtestenv}. The corresponding videos are provided in the supplemental file. On the TurtleBot2, MCB-K2 completed the navigation tasks in REnv1 and REnv2. In both scenarios, the robot traversed the constrained passages and reached the target without collision, including the multi-goal task in REnv2. In contrast, SCR failed in the same environments, where the robot became stuck near the narrow passage before reaching the target. The difference between the two methods was more pronounced on the Unitree Go2 legged platform. In REnv3, SCR collided with an obstacle in the cluttered indoor scene, whereas MCB-K2 adjusted its trajectory around the chairs and successfully reached the target. The Scout Mini in REnv4 followed the same pattern, with SCR colliding and MCB-K2 reaching the target without collision.
Additionally, we assessed the ability of the proposed method to handle dynamic obstacles in real-world scenes. As shown in Fig.~\ref{dynamicenv} and the supplemental video, the robots promptly adjusted their trajectories to avoid sudden obstacles and reached their designated goals. These results qualitatively showed that MCB-K2 could react to unexpected obstacle changes during real-world navigation.

\section{Conclusion}
This work revisits the common practice of resetting DRL navigation episodes after every collision and shows that this design choice can limit early-stage learning. The proposed multi-collision reset budget framework decouples local collision termination from global environment reset, allowing training to continue within the same scene while maintaining terminal targets for collision transitions. Simulation results show faster early-stage learning and improved navigation performance, while real-world trials on heterogeneous robot platforms further support the practical deployability of the learned policy in cluttered environments. These findings suggest that episode-boundary handling is an important design aspect for DRL-based navigation, alongside policy architecture and reward shaping.

\bibliographystyle{IEEEtran}
\bibliography{ref}

\end{document}